\documentclass{article}

\usepackage{arxiv}
\usepackage{cite}
\usepackage{amsmath,amssymb,amsfonts}
\usepackage{algorithmic}
\usepackage{graphicx}
\usepackage{textcomp}
\usepackage{xcolor}
\usepackage{url}
\graphicspath {{figure/}}
\usepackage{caption,subcaption}
\usepackage{url}
\def\BibTeX{{\rm B\kern-.05em{\sc i\kern-.025em b}\kern-.08em
    T\kern-.1667em\lower.7ex\hbox{E}\kern-.125emX}}

\usepackage{tikz}
\usepackage{textcomp}
\usepackage{hyperref}
\usepackage{lipsum}

\newcommand\copyrighttext{%
  \footnotesize \textcopyright 2020 IEEE. Personal use of this material is permitted.
  Permission from IEEE must be obtained for all other uses, in any current or future 
  media, including reprinting/republishing this material for advertising or promotional 
  purposes, creating new collective works, for resale or redistribution to servers or 
  lists, or reuse of any copyrighted component of this work in other works. \\
This paper was accepted for publication in the Fourth IEEE International Conference on Image Processing, Applications and Systems (IPAS 2020).
 }
\newcommand\copyrightnotice{%
\begin{tikzpicture}[remember picture,overlay]
\node[anchor=south,yshift=10pt] at (current page.south) {\fbox{\parbox{\dimexpr\textwidth-\fboxsep-\fboxrule\relax}{\copyrighttext}}};
\end{tikzpicture}%
}

\begin{document}
\copyrightnotice
\title{Are Adaptive Face Recognition Systems still Necessary? Experiments on the APE Dataset}

\author{Giulia Orrù$^1$, Marco Micheletto$^1$,Julian Fierrez$^2$, Gian Luca Marcialis$^1$
\\
$^1$\textit{University of Cagliari (DIEE, PRA Lab)}, Cagliari, Italy \\
$^2$\textit{Universidad Autonoma de Madrid (BiDA Lab)},Madrid, Spain \\
{\tt\small\{giulia.orru,marco.micheletto,marcialis\}@unica.it, julian.fierrez@uam.es}}

\maketitle

\begin{abstract}
In the last five years, deep learning methods, in particular CNN, have attracted considerable attention in the field of face-based recognition, achieving impressive results.
Despite this progress, it is not yet clear precisely to what extent deep features are able to follow all the intra-class variations that the face can present over time. In this paper we investigate the performance the performance improvement of face recognition systems by adopting self updating strategies of the face templates.
For that purpose, we evaluate the performance of a well-known deep-learning face representation, namely, FaceNet, on a dataset that we generated explicitly conceived to embed intra-class variations of users on a large time span of captures: the APhotoEveryday (APE) dataset\footnote{\url{https://github.com/PRALabBiometrics/APhotoEverydayDB}}. Moreover, we compare these deep features with handcrafted features extracted using the BSIF algorithm. In both cases, we evaluate various template update strategies, in order to detect the most useful for such kind of features. 
Experimental results show the effectiveness of ``optimized'' self-update methods with respect to systems without update or random selection of templates.
\end{abstract}

\section{Introduction}
In recent years, CNN-based Facial Recognition (FR) approaches have had a significant impact on both research and real applications. Contemporary deep learning approaches can achieve a close-to-zero error rate on popular datasets such as LFW.
On the other hand, intra-class variations such as poses, illuminations, expressions, and occlusions, still affect the performance of deep FR systems \cite{2018_IntelSys_Trends_Proenca,2017_IntelligentSystems_icb-rw}, which may benefit from pre-processing methods in order to normalize those variability factors \cite{2011_QualityBio_FAlonso}. 
The excellent performance and robustness of deep FR systems might suggest that the performance of deep templates is stable over time, but some works have showed that it is not yet clear to what extent face representations are able to keep high performance over large time lapses between enrollment and testing data or large variations of other factors. 

In the recent past, several ``adaptive'' biometric systems have been proposed to deal with changing context and variability factors \cite{Fierrez-Aguilar2005_BayesianAdaptation,Fierrez-Aguilar2005_AdaptedMultimodal,Roli2008,2018_INFFUS_MCSreview2_Fierrez}. As described in \cite{2018_INFFUS_MCSreview2_Fierrez}, any component of a biometric system is subject to time adaptation. Here we focus in updating the biometric templates \cite{2019_IETB_TemplateUpdate_Tolosana}. The concept of template adaptation is referred to the generation of novel templates in time that can replace or can be coupled with existing ones \cite{ULUDAG20041533,galbally13PONEagingSignature}. This is done automatically by exploiting samples captured during normal system operations.
As highlighted by \cite{ORRU2020107121}, template update improves the performance of facial recognition systems, including those based on deep-learning techniques.

In practice, template update (also referred to as self update from now on) has various advantages with respect to the network's re-tuning to track variations in time: (1) templates can be easily stored in the user's own device with an irrelevant memory waste, (2) no need for accessing the original data, namely the users' images, in order to retrain the network, and (3) this avoid privacy issues.

The evaluations carried out so far in related works, however, usually consider datasets collected over a short time span. Therefore, these datasets do not allow to analyze properly how the typical intra-class variations of aging are followed by template update methods.

The goal of this work is to perform an evaluation of deep features for Face Recognition over time with and without template update by using a dataset that we generated explicitly conceived to contain many intra-class variations, long acquisition time span, and temporal information: the APhotoEveryDay (APE) dataset. In fact, exploiting the temporal information allows us to carry out an analysis that realistically simulates the normal adaptive facial recognition system operation.

In our experiments we analyze two FaceNet models in comparison with a BSIF-based handcrafted approach. Both basic and more recent template update strategies are investigated, in order to detect the most useful for such kind of features. 
Experimental results show that template update is helful to improve the evaluated face recognition systems over time. 

The paper is organized as follows. Section \ref{featurerep} summarizes the evolution of feature representations and the weaknesses of the approaches used in the last decades. Section \ref{methods} describes the adaptive methods implemented and evaluated. Section \ref{materials} describe the face models and dataset used in the experimental evaluation. Section \ref{result} reports the experimental results and discussions. Section \ref{conclus} concludes the paper.

\section{Feature representation in Face Recognition}
\label{featurerep}
\begin{table*}[ht]
\centering
\resizebox{\textwidth}{!}{%
\begin{tabular}{|l|c|c|c|c|}
\hline
\multicolumn{1}{|c|}{\textbf{Feature represent.}} & \textbf{Holistic learning}                                                              & \textbf{Local handcraft}                                                           & \textbf{Shallow learning}                                                                  & \textbf{Deep learning}                                                                                                   \\ \hline
\textbf{Introduction (years)}                         & 1990s                                                                                   & 2000s                                                                              & 2010s                                                                                      & 2012/2014                                                                                                                \\ \hline
\textbf{Issues}                                       & \begin{tabular}[c]{@{}c@{}}Fail to address\\ uncontrolled\\ facial changes.\end{tabular} & \begin{tabular}[c]{@{}c@{}}Lack of distinctiveness \\ and compactness.\end{tabular} & \begin{tabular}[c]{@{}c@{}}Fail to address \\ facial appearance\\  variations.\end{tabular} & \begin{tabular}[c]{@{}c@{}}Black-box. Requires time,\\ powerful hardware and a large number \\ of images for training.\end{tabular} \\ \hline
\end{tabular}}
\caption{\label{featureevolution}Feature representation evolution in Face Recognition systems.}
\end{table*}

Over the past years, many methods have been developed to address the issues and challenges of facial recognition systems. In particular, the non-stationarity of the face appearance, which presents many intra-class variations such as aging, lighting, pose, and occlusions, is often addressed by implementing adaptive systems.
These biometric systems ``adapt'' continuously the gallery templates to the variations of the input data, without the need of human intervention.

The performance of adaptive systems depends on the expressive power of the face representation methods adopted. These can be grouped into four broad categories (Table \ref{featureevolution}) that have been developed in different periods of the past thirty years \cite{Wang2018DeepFR}. In holistic approaches, widespread since the 90s, the complete face region is taken into account as input data and the face information is represented by a small number of features \cite{eigenFace,pcaface}. The main issue of this approach is that these methods fail to address uncontrolled facial changes.

To solve this problem, local features of faces were proposed to select a number of features to uniquely identify individuals, the so-called local handcrafted methods, in the early 2000s. Among them, Local Binary Patterns (LBPs) \cite{1017623,1717463}, Gabor \cite{999679}, BSIF \cite{BSIF}, and their variants \cite{1541333} exhibit as limitation the lack of compactness and in some cases a low distinctiveness among individuals. 

The problem of the low compactness level was faced at the beginning of the 2010s; in particular, shallow learning methods were introduced. These methods, also known as learning-based local descriptors, use unsupervised learning methods to encode the local microstructures of the face into discriminative codes \cite{5539992}.
Unfortunately, shallow learning methods cannot handle complex nonlinear facial appearance variations such as self-occlusions and pose variations \cite{Ding:2016:CSP:2885506.2845089}.

After 2012, deep learning methods began to spread and achieved state-of-the-art results in many problems. These methods use a cascade of processing layers to learn representations of data with multiple levels of abstraction.
From 2014 also the Face Recognition community adopted deep learning methods, such as convolutional neural networks, for facial feature extraction and transformation. Among others, DeepFace \cite{deepface}, FaceNet \cite{facenet}, VGGFace \cite{Parkhi15} and VGGFace2 \cite{DBLP:journals/corr/abs-1710-08092}. Some of them achieved state-of-the-art performance on the most challenging datasets known, such as LFW \cite{LFWTech}, IJB-A \cite{ijba}, IJB-B  \cite{ijbb}, etc.

Although the good results obtained, deep learning approaches work as “black-box” feature extractors from face images \cite{8244393}. For this reason it is not yet clear how these systems can handle data that present significant time changes. 
Moreover, the datasets mentioned above do not contain temporal information in order to build a realistic time sequence of facial images of the same people across months or years.

As stated in the Introduction, our main aim is to understand if the compact and powerful representation obtained through deep learning methods is able to work in situations of long-term use in which the temporal variability of the biometric data is more accentuated.

\section{Adaptive Methods for Template Update}
\label{methods}

First of all, we implemented standard facial recognition as a sort of ``ground truth'', since our main claim is to clarify to what extent self update approaches are helpful over top-performing deep-learning based face recognition. In particular, we tested the ``traditional'' self-update system \cite{Rattani2009}, the classification-selection method based on risk minimization \cite{RattaniDual} and two methods of classification-selection with limited number of templates per user, based on K-means \cite{1247} and on the semi-supervised application of RANDOM editing methods just to verify if the selection performed by the two methods above is significant or not.
In fact, it is possible to categorize biometric adaptive systems into two categories:
\begin{itemize}
    \item The traditional self-update system that only classifies input samples and adds them to the gallery if they meet the genuinity requirements.
    \item Classification/selection approach in which the selection phase allows to filter the redundant information \cite{ape,ORRU2020107121}.
\end{itemize}

Traditional self-update estimates an updated threshold through the gallery and, when a \textit{batch} \cite{Rattani2013AMD} is available for a certain claimed user, the distances between each sample and the user's template(s) are computed. Input samples whose distance is less than the update threshold are added into the user's gallery.
Classification/selection algorithms add another phase where the best samples for the gallery are chosen.

In particular, Rattani \textit{et al.} introduced a classification/selection system based on harmonic functions and a risk minimization technique \cite{RattaniDual}.
In \cite{1247} the authors present a method that keeps the number of templates constant at each iteration by setting the maximum number of images per user, namely, $p$, in the selection phase. This is obtained by deriving the centroid of the samples of a given subject through the K-means algorithm, and then selecting the $p$ closest samples to that centroid. 

Finally, the RANDOM method simply selects $p$ pseudo-labeled samples of each user. This method allows us to simulate what may happen by keeping a human in the loop for selecting the best templates to update. Indeed, the selection of a human supervisor is unpredictable and changes depending on the individual involved.

\section{Face Models and Dataset}
\label{materials}

\subsection{Face Models: FaceNet and BSIF}
We use two FaceNet models \cite{facenet} and handcrafted Binarized Statistical Image Features (BSIFs) \cite{BSIF}. 

FaceNet uses a deep convolutional network to optimize the representation of an individual's face through a 128 bytes feature vector. In particular, faces are mapped into a 128-dimensional Euclidean space in which distances directly correspond to a measure of facial similarity.
The FaceNet architecture is based on the triplet loss function \cite{Weinberger:2009:DML:1577069.1577078}: the feature vectors related to the same individual have small distances, while those of distinct people have large distances.
To evaluate the effectiveness of the FaceNet representation, we used an open-source implementation based on TensorFlow \cite{facenetimpl} trained on the model 20170512-110547 and on the model 20180408-102900, whose characteristics are shown in Table \ref{tab:detmodel}. The first model has been trained on the MS-Celeb-1M dataset \cite{guo2016msceleb} and represents the face with a 128 byte feature vector. The second model has been trained on the CASIA-WebFace dataset \cite{Yi2014LearningFR} and uses a 512 byte feature vector.

\begin{table}[ht]
    \centering
    \begin{tabular}{l|c|c}
    \textbf{Model} & \textbf{20170512-110547} & \textbf{20180408-102900}\\
    \textbf{Dataset} & MS-Celeb-1M & CASIA-WebFace	\\
    \textbf{Embedding size}  & 128D & 512D\\
    \textbf{Image stand.}	& Per image & Fixed\\
    \textbf{Data augment.}	&	Random crop/flip & Random flip\\
    \textbf{Optimizer }&	RMSProp & Adam\\
    \end{tabular}
    \caption{FaceNet models details.}
    \label{tab:detmodel}
\end{table}

\subsection{APE Dataset}

\begin{figure}[htb]
\centering
\begin{subfigure}[b]{1\linewidth}
\centering
    \includegraphics[width=0.5\linewidth]{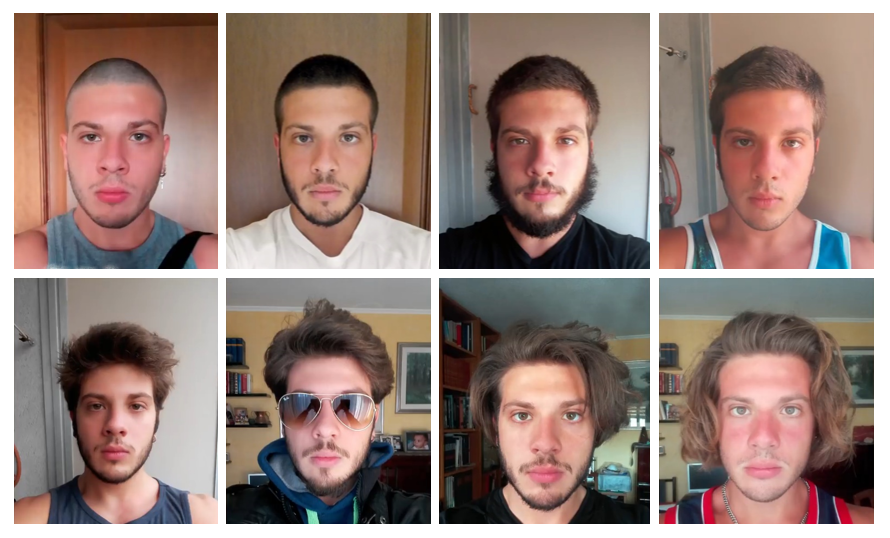}
      \end{subfigure}%
      \\
\begin{subfigure}[b]{1\linewidth}
 \centering
   \includegraphics[width=0.5\linewidth]{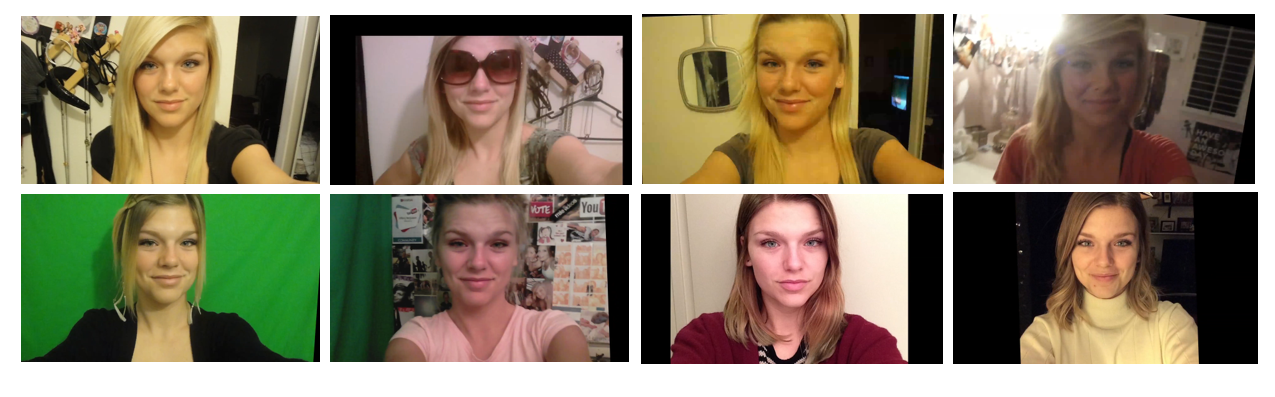}
      \end{subfigure}%
    \caption{\label{example}Examples of face images from the APE dataset exhibiting many variations over time.}
\end{figure}

Using FaceNet we extracted the feature vectors related to the “APhotoEveryday” (APE) dataset \cite{ape} that contains the faces of 46 individuals with large appearance variations across time.
The APE dataset was acquired by the University of Cagliari and consists of facial images extracted from YouTube Daily Photo Projects.
Faces were captured in the frontal pose and most of these have a controlled expression. The images of each user are labelled with a number that indicates the temporal progression of the sequence.
The APE dataset images are characterized by many temporal variations of the facial appearance thanks to the Daily Photo Project\footnote{See for example \url{https://www.youtube.com/watch?v=RBPYDIzEbYk}}.
The number of images per user varies between 92 and 3892 and the acquisition time varies between less than one year and twelve years. Fig. \ref{example} shows some facial images from the APE dataset.

\section{Experiments}
\label{result}

\subsection{Experimental Protocol}

The experimental protocol was designed in order to exploit the full potential of the dataset, in particular the high number of samples per user and the temporal information associated with them. This is summarized as follows:
\begin{itemize}
    \item The dataset was subdivided into ten parts maintaining the sequence time progression: the first partition is the initial gallery and is composed of the first $p=5$ images of each user; the remaining nine parts are composed of $\frac{\# samples - p}{\# adaptation sets}$ samples for each user.
    \item The operational points FAR and FRR are calculated and the updating threshold was estimated at FAR=$1\%$.
    \item The adaptation sets were used to simulate the periodic sets of batches collected during the system’s operations individually processed for updating the users’ galleries.
    \item The test set was used to evaluate the system’s performance from the state made up of the initial galleries, to the updated gallery after each updating cycle. In order to better simulate a real application and exploit time information we used the $(i+1)^{th}$ batch as test set of the $i^{th}$ batch as suggested in \cite{5949222}.
\end{itemize}

\subsection{Results}

\begin{figure}[tb]
\centering
  \begin{subfigure}[b]{0.49\linewidth}
   \centering \includegraphics[width=1\textwidth]{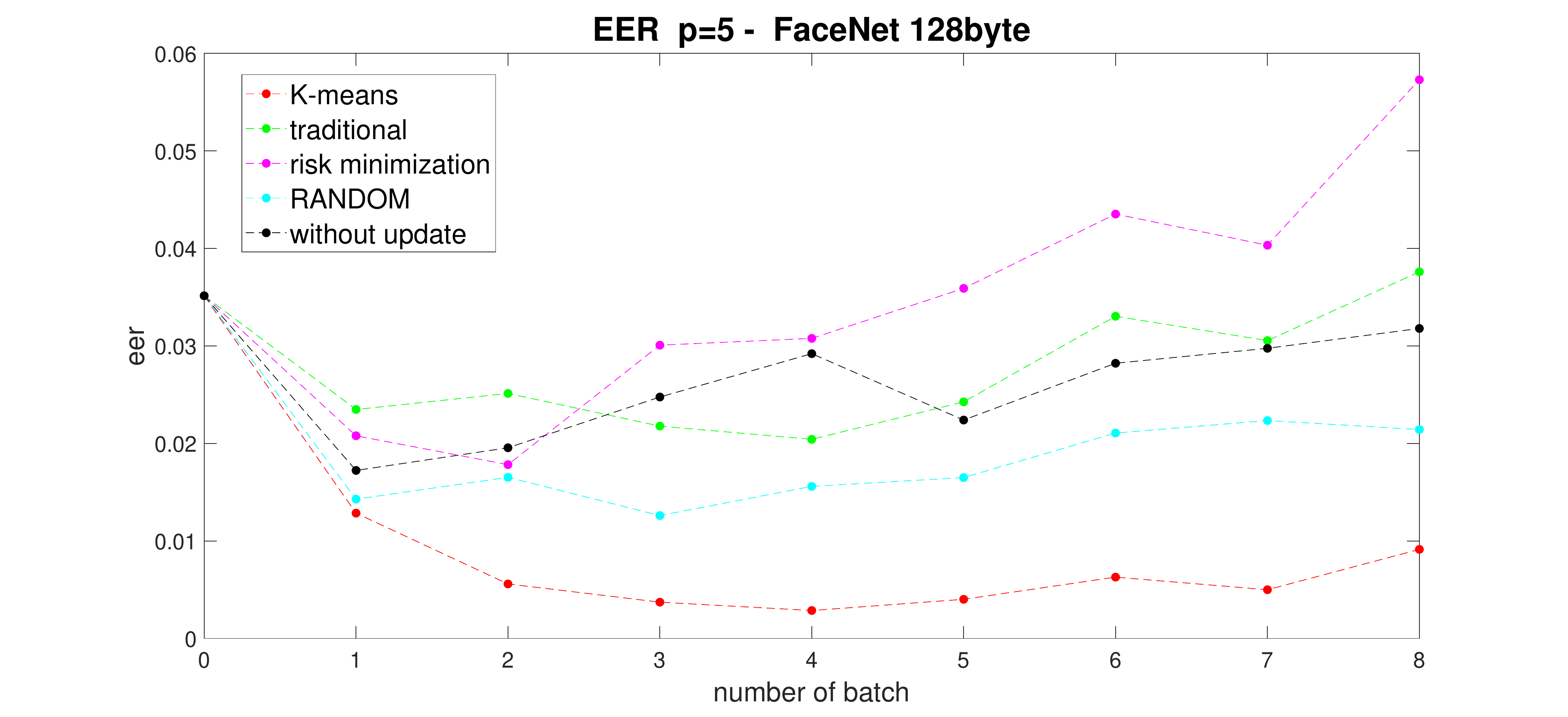}
    \subcaption{EER - FaceNet 128 byte - APE dataset. For x=$i$: update on batch $i$, performance on batch $i+1$.}
  \end{subfigure}
\begin{subfigure}[b]{0.49\linewidth}
    \centering\includegraphics[width=1\textwidth]{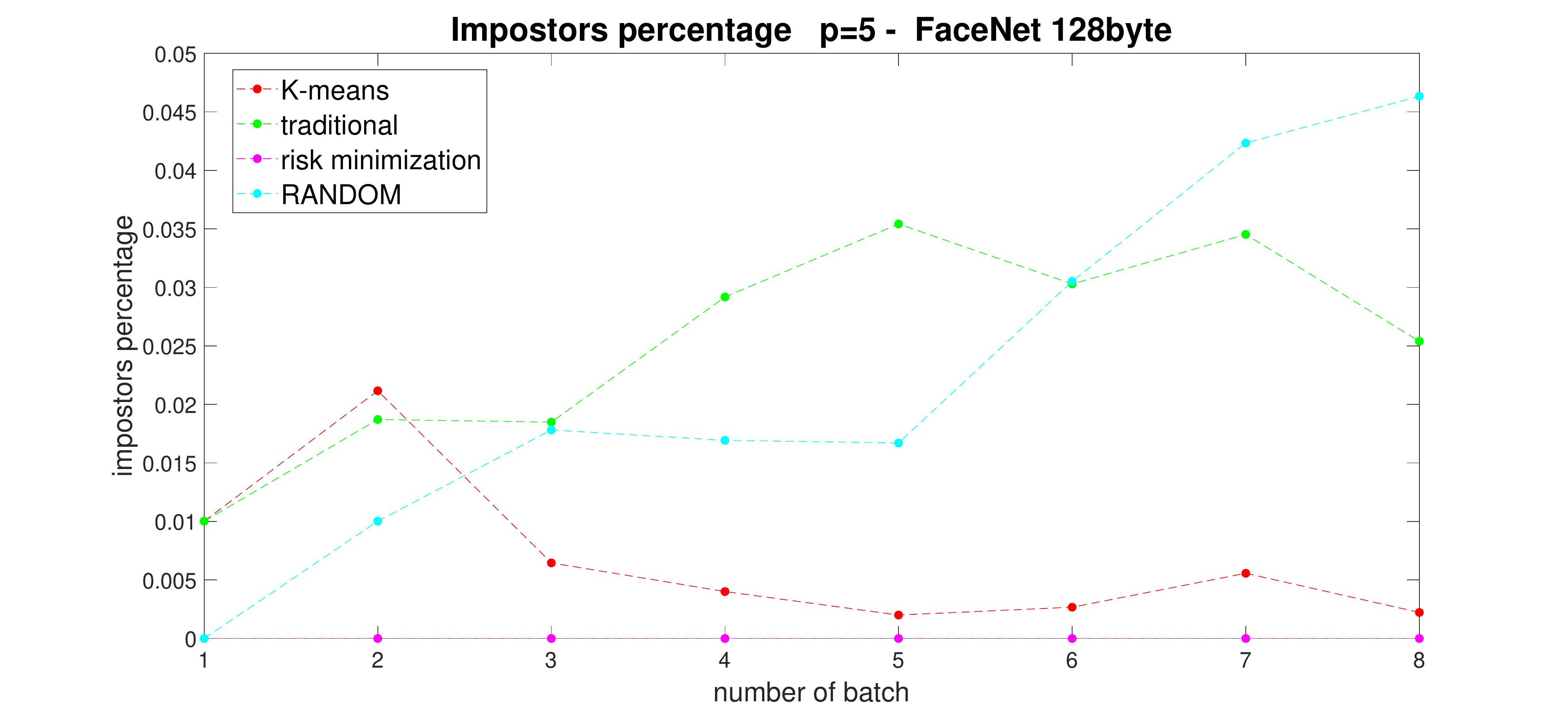}
    \subcaption{Impostors percentage - FaceNet 128 byte - APE dataset.\newline}
  \end{subfigure}%
      \caption{\label{res128}EER and percentage of impostors for different template update methods with $p$=5 for FaceNet 128D (APE dataset).}
\end{figure}

\begin{figure}[tb]
\centering
  \begin{subfigure}[b]{0.49\linewidth}
   \centering \includegraphics[width=1\textwidth]{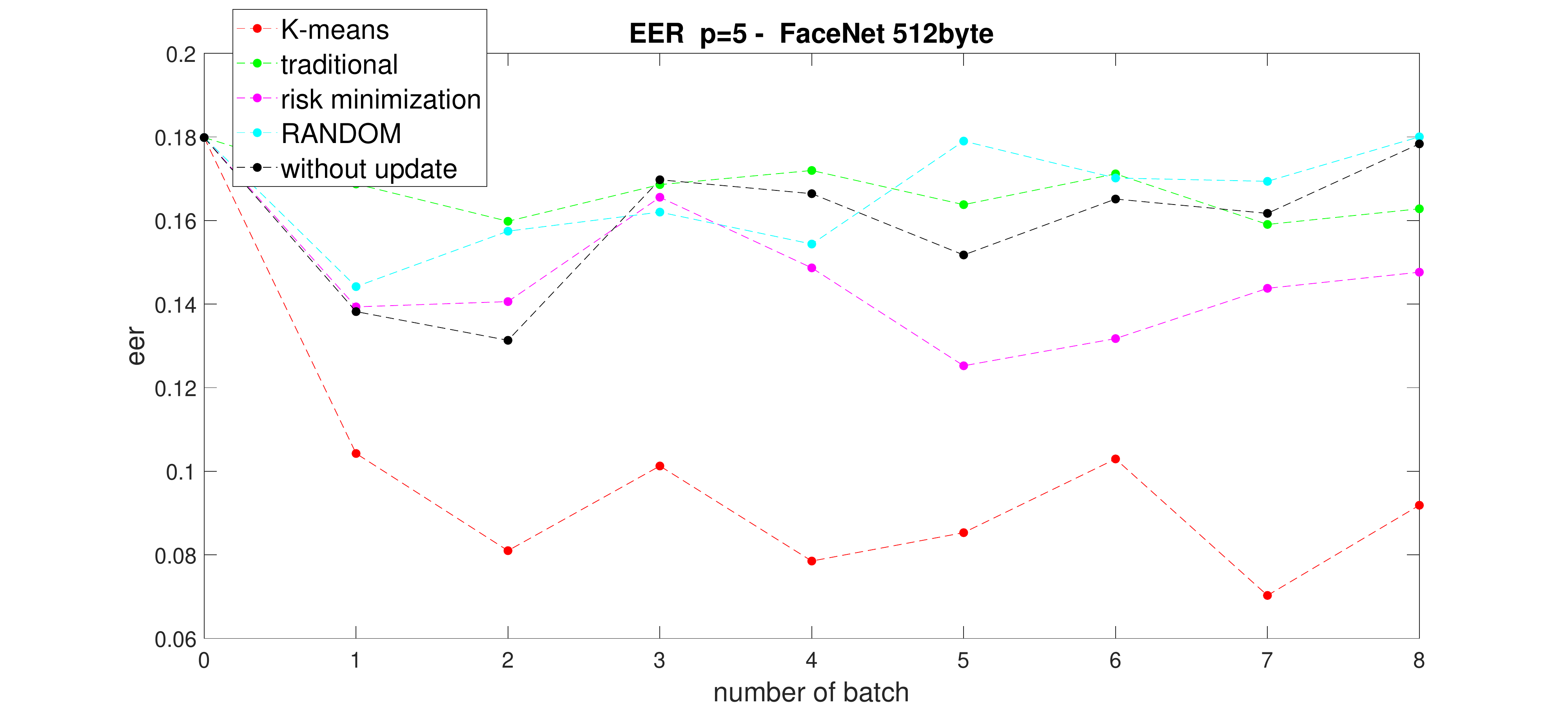}
    \subcaption{EER - FaceNet 512 byte - APE dataset. For x=$i$: update on batch $i$, performance on batch $i+1$.}
  \end{subfigure}
\begin{subfigure}[b]{0.49\linewidth}
    \centering\includegraphics[width=1\textwidth]{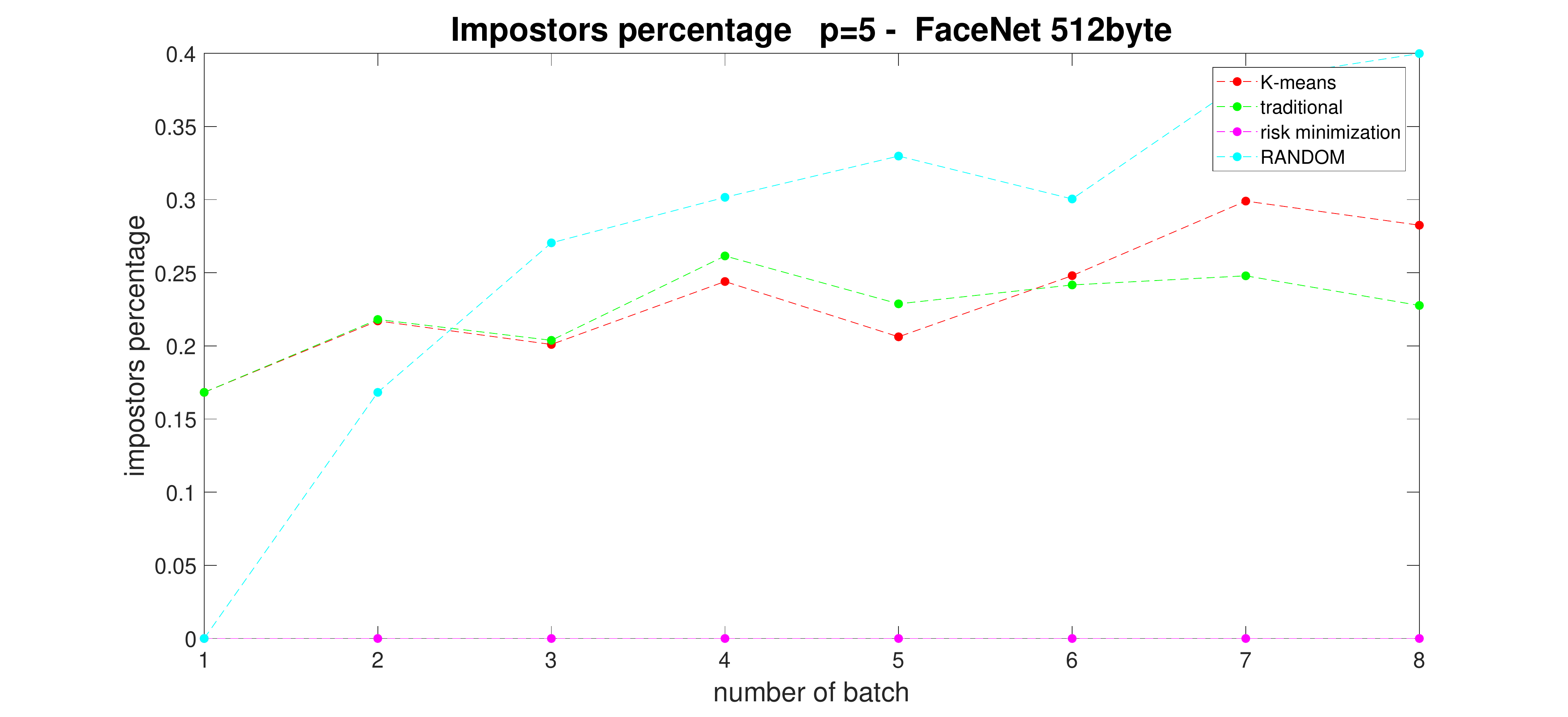}
    \subcaption{Impostors percentage - FaceNet 512 byte - APE dataset.\newline}
  \end{subfigure}%
      \caption{\label{res512}EER and percentage of impostors for different template update methods with $p$=5 for FaceNet 512D (APE dataset).}
\end{figure}

\begin{figure*}[tb]
\centering
\includegraphics[width=0.49\textwidth]{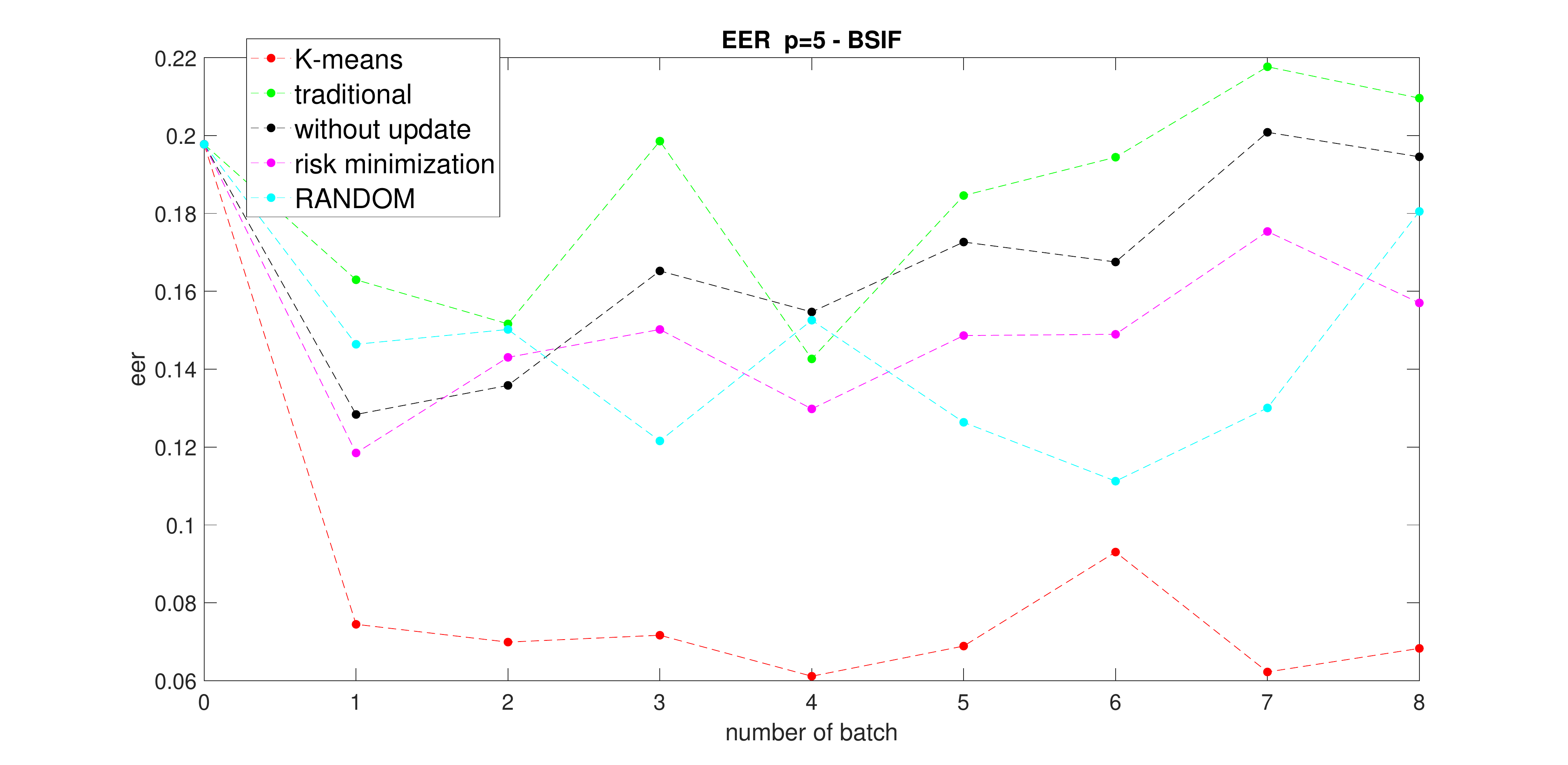}
\includegraphics[width=0.49\textwidth]{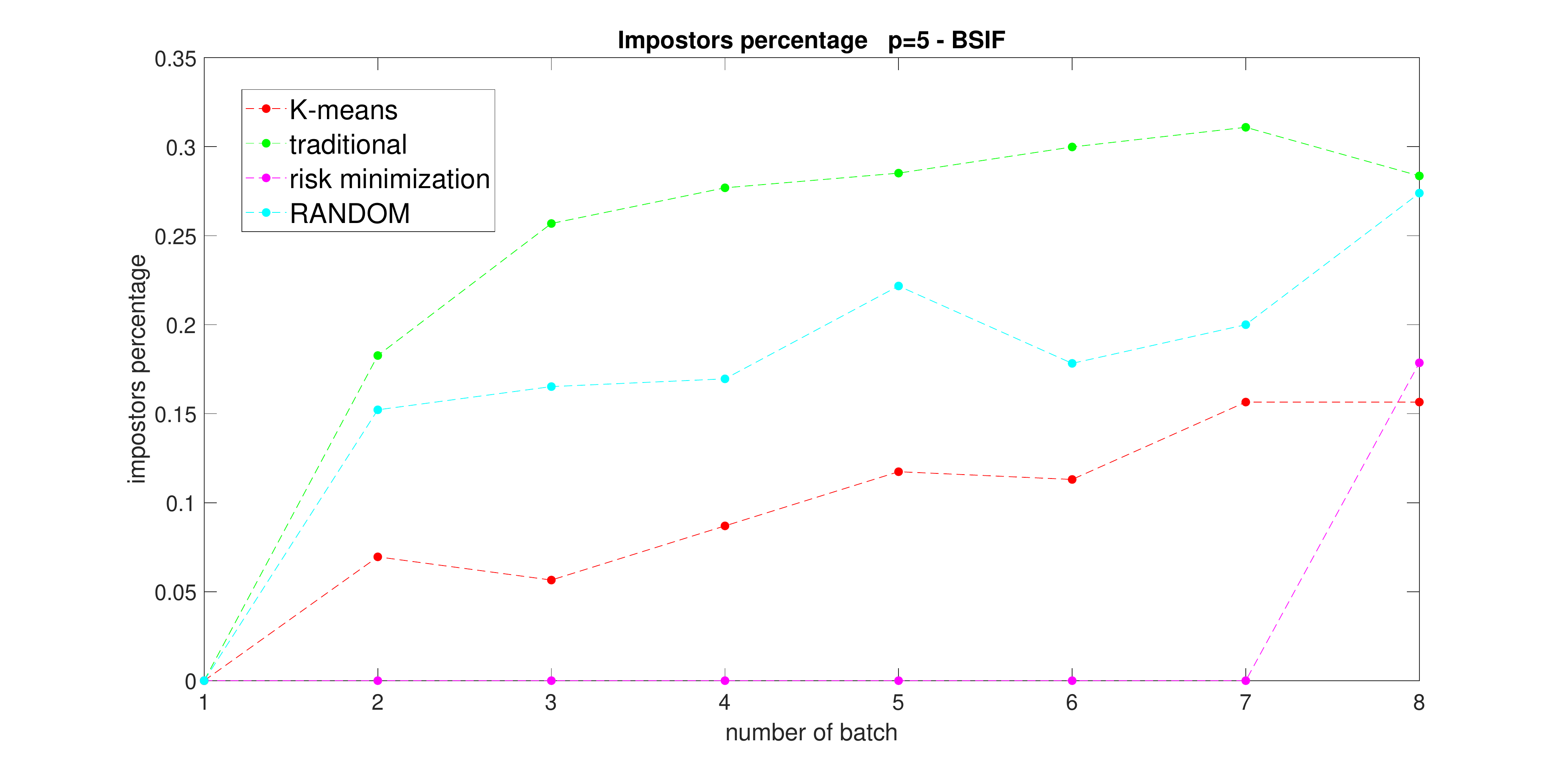}
\caption{EER and percentage impostors comparison among the SOTA methods with p=5 for BSIF feature vectors. On the x axis is shown the number of the batch and on the y axis the performance index.}\label{resbsif}
\end{figure*}

The traditional self-update \cite{Rattani2009},  the  method  based  on  risk-minimization \cite{RattaniDual}, the K-means \cite{1247}, and the RANDOM classification/selection method were implemented and tested to evaluate the performance using feature vectors extracted with FaceNet 128 dimension embeddings (Fig. \ref{res128}), FaceNet 512 dimension embeddings (Fig. \ref{res512}) and with the handcrafted BSIFs (Fig. \ref{resbsif}).

The most remarkable result emerging from Fig. \ref{res128}(a) and Fig. \ref{res512}(a) is that for deep features, despite their compactness and representativeness, classification/selection template update methods improve their performance compared to a solution without self-update (black line) or to the random selection of templates, which can be assimilated to the update supervised by a human operator. In particular, for both FaceNet models, the K-means is the best algorithm. In fact, this classification/selection method, as demonstrated in \cite{ORRU2020107121}, allows at the same time to keep the system error low and a limited number of templates in memory. The result is low computational complexity and stable performance over time.

The performance of the traditional self-update (i.e., adding templates to the system without selection), is worse in terms of accuracy than the system without updating, in addition to having high requirements for template storage. This is probably due to the fact that there is no filtering of introduced impostors. This drift was observed in early publications too \cite{Roli2006} and it is shown by the increasing percentage of impostors present for all models analyzed (Figs. \ref{res128}b, \ref{res512}b, \ref{resbsif}b). 

\begin{figure*}[tbp]
\centering
\includegraphics[width=0.49\textwidth]{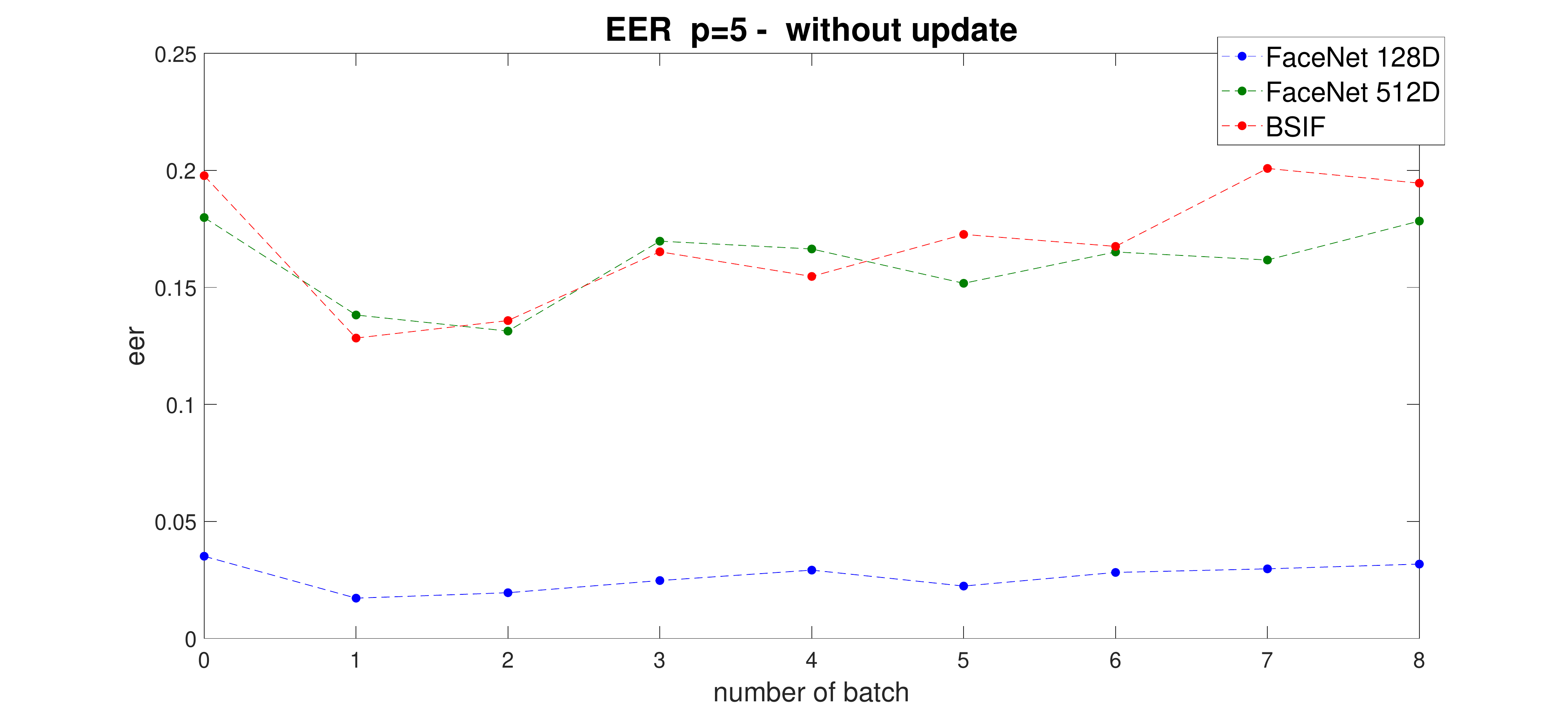}
\includegraphics[width=0.49\textwidth]{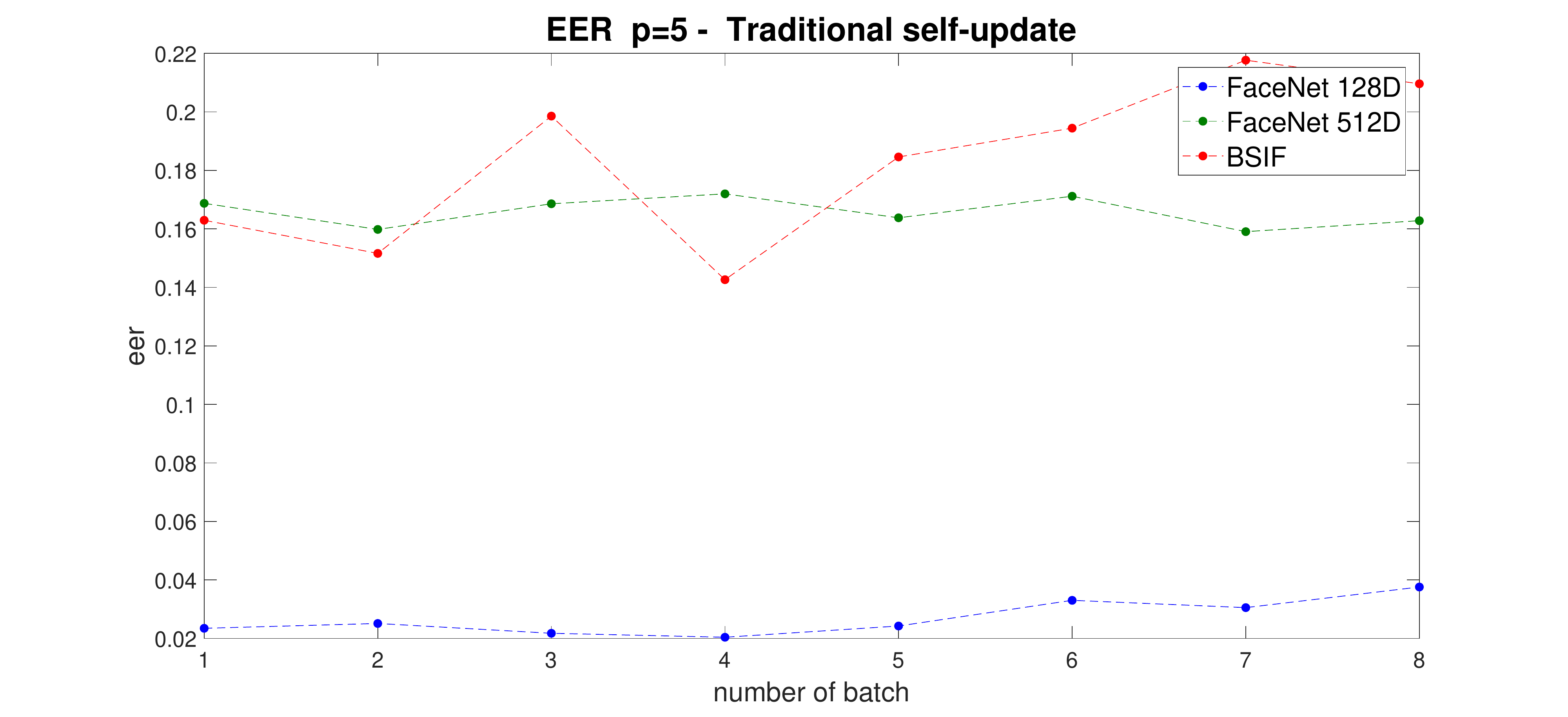}
\includegraphics[width=0.49\linewidth]{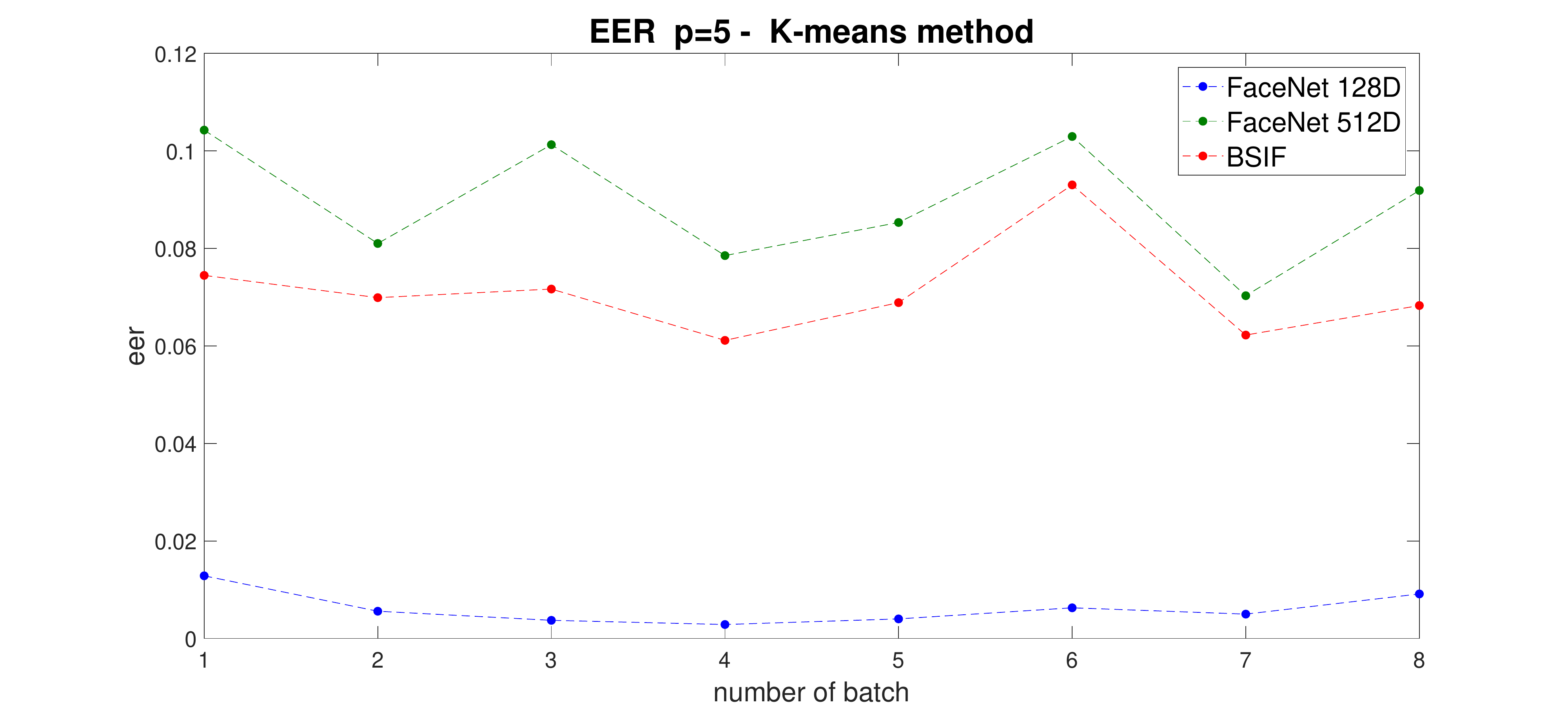}
\caption{EER comparison between the different feature vectors tested: without update, traditional update, kmeans-based update.}\label{confrontofeer}
\end{figure*}

Fig. \ref{confrontofeer} shows the performance of the three feature extraction approaches for facial recognition without updating, with traditional self-update and self update by classification/selection (K-means).
These results highlight how the system based on 128D FaceNet performs better than the one based on 512D FaceNet and BSIFs. The better performance of the 128D model compared to the 512D model is probably due to the data used for training the neural network.

Table \ref{tableeer} supports this claim; the EER calculated on the initial gallery and on the first batch and the averages and standard deviations along the remaining batches (with update on the $i$ batch and performance on the $i+1$ batch for $i>1$) are reported comparing the Face Recognition performance without update with the traditional self-update and the K-means classification-selection approach. The reported results show that the template update driven by K-means has a very positive effect on both global performance and system stability. In particular, for FaceNet128D the error goes from 2.54\% for the method without updating to 0.62\% for the K-means method.
These results demonstrate that, depending on the application context, the use of template update and selection allows a substantial improvement in the performance of Face Recognition systems resulting more adequate than the intervention of a human operator.

\begin{table*}[tbp]

\centering
\begin{tabular}{|c|c|c|c|c|}
\hline
\multicolumn{5}{|c|}{\textbf{EER}}                                                                                                                                                  \\ \hline
                                       & \textbf{Initial Gallery} & \textbf{Without update}             & \textbf{Traditional self-update}    & \textbf{K-means}                    \\ \hline
 {\textbf{FaceNet 128D}} &  {3.51\%}  &  {2.54\%  $\pm$ (0.52)}   &  {2.70\%  $\pm$ (0.60)} & {\textbf{0.62\%  $\pm$ (0.33)}} \\ \hline
 {\textbf{FaceNet 512D}} &  {17.99\%}  &  {15.78\%  $\pm$ (1.62)} &  {16.58\%  $\pm$ (0.50)} &  {8.94\%  $\pm$ (1.27\%)} \\ \hline
 {\textbf{BSIF}}         &  {19.78\%}  &  {16.50\%  $\pm$ (2.54)}  &  {18.28\%  $\pm$ (2.75)} &  {7.12\%  $\pm$ (1.00)} \\ \hline
\end{tabular}

      \caption{\label{tableeer}EER comparison among the different feature vectors tested: the table shows the EER calculated by training on the initial gallery and testing on the first batch and the averages and standard deviations calculated along the remaining batches (with update on the $i$ batch and performance on the $i+1$ batch for $i>1$).}

\end{table*}

\section{Conclusions}
\label{conclus}
In this paper we analyzed the template representativeness of deep-learning features, in particular those extracted with two models of FaceNet, to understand if adaptive systems are useful in modern Face Recognition systems. In order to make a comparison between hand-crafted and deep learning features we have also analyzed the system performance with handcrafted BSIF features. To evaluate the performance across a large time lapse between enrollment and testing, and to simulate a challenging scenario including drastic appearance changes, the experiments were conducted on the APE dataset.
The evidence from this study shows the benefits of update methods with classification/selection in situations where the face appearance presents many intra-class variations \cite{2018_TIFS_SoftWildAnno_Sosa}. In fact, the use of ``optimized'' template update allows a substantial improvement in the performance compared to systems without updating or systems that keep the human in the loop, simulated with a random selection of templates.
It is therefore possible to say that adaptive systems are useful on top of modern deep face recognition, at least in the scenarios considered here in our APE face image dataset (with time spans of months to years between enrollment and testing).
\section*{Acknowledgements}
This work is supported by the Italian Ministry of Education, University and Research (MIUR) within the PRIN2017 - BullyBuster - A framework for bullying and cyberbullying action detection by computer vision and artificial intelligence methods and algorithms (CUP: F74I19000370001), and by the Spanish project BIBECA (RTI2018-101248-B-I00 from MINECO/FEDER).

\bibliographystyle{IEEEtran}

\bibliography{lniguide}

\end{document}